\documentclass[conference]{IEEEtran}
\IEEEoverridecommandlockouts
\usepackage{cite}
\usepackage{amsmath,amssymb,amsfonts}
\usepackage{algorithmic}
\usepackage{graphicx}
\usepackage{todonotes}
\usepackage{textcomp}
\usepackage{xcolor}
\def\BibTeX{{\rm B\kern-.05em{\sc i\kern-.025em b}\kern-.08em
    T\kern-.1667em\lower.7ex\hbox{E}\kern-.125emX}}

\usepackage[hyphens]{url}
\usepackage{colortbl}

\begin{document}

\title{Ensemble Decision Systems\\ for General Video Game Playing}

\author{\IEEEauthorblockN{Damien Anderson, Philip Rodgers and John Levine}
\IEEEauthorblockA{\textit{Computer and Information Sciences}\\
\textit{University of Strathclyde}\\
Glasgow, United Kingdom \\
\{damien.anderson, philip.rodgers, john.levine\}@strath.ac.uk}
\and
\IEEEauthorblockN{Cristina Guerrero-Romero and Diego Perez-Liebana}
\IEEEauthorblockA{
\textit{School of Electronic Engineering and Computer Science}\\
\textit{Queen Mary University of London}\\
London, United Kingdom \\
\{c.guerreroromero, diego.perez\}@qmul.ac.uk}
}

\IEEEpubid{\begin{minipage}{\textwidth}\ \\[12pt]
978-1-7281-1884-0/19/\$31.00 \copyright 2019 IEEE
\end{minipage}}
\maketitle

\begin{abstract}
Ensemble Decision Systems offer a unique form of decision making that allows a collection of algorithms to reason together about a problem. Each individual algorithm has its own inherent strengths and weaknesses, and often it is difficult to overcome the weaknesses, while retaining the strengths. Instead of altering the properties of the algorithm, the Ensemble Decision System augments the performance with other algorithms that have complementing strengths. This work outlines different options for building an Ensemble Decision System as well as providing analysis on its performance compared to the individual components of the system with interesting results, showing an increase in the generality of the algorithms without significantly impeding performance.
\end{abstract}


\section{Introduction}

When developing agents to play a game, characteristics of the game under consideration can be included in the logic to guide them to the objective and play optimally, but that agent will not be able to perform well in other games or when a rule is updated. General Video Game Playing (GVGP) aims to tackle the challenge of building agents capable of performing well in different games, so these agents' value functions need to be as general as possible.

Most of the heuristics present in planning algorithms developed for GVGP  focuses on specific goals, which usually are winning or guiding the agent following the maximization of the score. Others also include information about the proximity of different elements of the game, which is combined with the previous ones. The agents follow these goals and they perform well in certain types of games, while the reward structure is well defined, or the goals are in a reachable distance. However, what happens when the game is built in a way where the reward structure is not clear or is designed to guide the agents away from the optimal solution? What happens in games with large maps where the agent needs to move in a particular path to reach the goal? The agents who focus on just the goal and score will have a poor performance; being unable to solve the games. A solution would be building an agent with new goals that overcomes the weakness of that heuristics (e.g. focused on exploring the level). Nevertheless, even when this new agent could solve the games that the other ones were unable to, it may become worse at games that it previously did well at, which is not the outcome we are looking for.

The approach taken by some authors to tackle this problem is using systems that combine different agents based on the type of game they believe they are facing, but their performance relies entirely on the correct prediction of the game. The solution we present in this paper is using an Ensemble Decision System (EDS), which is built to make the most of different agents by focusing on their strengths instead of weakness. We carried out an experiment where we ran a series of EDS with different configurations and compared their performance in contrast with known sample agents provided in the General Video Game AI framework. The results show the flexibility of this approach and open an interesting GVGP line of research to develop its full potential.

\section{Background}
The field of General Video Game Playing (GVGP) began in $2013$ as a way of promoting interest towards developing game playing agents that could play a wide variety of video games \cite{levine2013general}. The initial interest in this field originally sprang from work in General Game Playing and a desire to explore game playing agents in real-time situations \cite{genesereth2005general}. 

The GVGP field is supported through an online competition known as the General Video Game AI competition\footnote{http://www.gvgai.net/} (GVGAI) where entrants can submit GVGP agents. The GVGAI competition offers a framework that provides a large library of games, up to $122$ currently. Further games can be quickly created and added to the library through the Video Game Description Language (VGDL) that simplifies the game creation process \cite{ebner2013towards}. The competition also offers a variety of tracks for different types of agents. These tracks focus on two main areas, planning and learning \cite{perez20162014} \cite{liu2017single}. The planning track offers both a single player and multi-player variation, which provide agents with a forward model in order to facilitate search-based algorithms. Agents can use the forward model to search through and evaluate future states of the game. The learning track, on the other hand, removes the forward model and replaces it with a training period before official evaluation occurs which allows agents to develop a policy for playing the game. The work carried out for this research focuses on the single player planning track.

Over the years the competition has received a large number of entrants \cite{perez2018general}, though only a few have attempted to incorporate multiple types of algorithms into one system. These agents typically take on a portfolio approach, however, determining the type of game being played and applying a single algorithm to that game. One such example is \emph{YOLOBOT}, which uses the game dynamics observed to determine whether a game is deterministic or stochastic, running, respectively, a heuristic Best First Search or Monte Carlo Tree Search (MCTS) \cite{joppen2018informed}. Another agent, \textit{Return42}, uses a similar approach but instead uses an A* algorithm for stochastic games, and random walks for deterministic games \cite{ashlock2017general}. These approaches have enjoyed a great deal of success in the GVGAI competition, yet, rely largely on the correct identification of the type of game to fully leverage the strength of the algorithms. Mendes \textit{et al.} used a portfolio hyper-agent approach that predicted which of the seven controllers they included should be used based on the features present in the game; outperforming the winners of the $2014$ and $2015$ competitions \cite{mendes2016hyper}.

The concept of tackling complex problems with an array of algorithms has found success in other areas of AI research. Most notably, Google Deepmind's \emph{AlphaGo} makes use of multiple systems to create complex behaviour that is capable of defeating professional human players at the game of Go \cite{silver2017mastering}. By using a combination of neural networks and tree search algorithms, the system is capable of succeeding where no single algorithm has been able to so far. Similarly, an Ensemble Decision System (EDS) has achieved a world record for an AI playing the game of \textit{Ms. Pacman}. The EDS achieves this by allowing complex behaviour to emerge from the combining of simple algorithms that focus on specific tasks, such as collecting pills or dodging ghosts \cite{rodgers2018ensemble}.
%% I swaped the order of these two paragraphs (up and down) because I think they relate better like this

One of the main challenges to overcome is the notion of games, or environments, that are purposefully designed to lead agents away from an optimal outcome \cite{anderson2018deceptive}. Perez \textit{et al.} took a look at the robustness of agents to disruptive changes in the environment, such as a forward model that would return incorrect information, or a system that would occasionally apply a random action instead of the agent's intended action for a given state \cite{perez2016analyzing}.
%This last paragraph doesn't relate well to the rest of the paper. Might need to add something to show why it is important

\section{Controllers}\label{Controllers}

In this section we present, and briefly describe, the algorithms and heuristics that have been either included in the EDS implemented (Section~\ref{sec:EDS}) or executed for performance comparison.

\subsection{Algorithms}
The controllers used in these experiments belong to the sample pool of the GVGAI framework.

\subsubsection{sampleRandom}
The action is chosen randomly between the options available. %There is no decision making process or reasoning in the logic.

\subsubsection{One Step Look Ahead (OSLA)}
It estimates the reward gained for each of the possible actions on the next step of the game and chooses the action that returns the highest value. The \textit{sampleOneStepLookAhead} agent uses by default the \textit{SimpleStateHeuristic}, provided in the framework.

\subsubsection{Open-Loop Monte-Carlo Tree Search (OLMCTS)}
It is a variant of Monte-Carlo Tree Search (MCTS) \cite{browne2012survey} that is designed to work better in stochastic environments. It uses the forward model to reevaluate the actions instead of keeping the states of the game in the nodes of the tree. The \textit{sampleMCTS} version provided in the framework has a rollout length of $10$ and a C-value of $\sqrt{2}$. The default value function takes into consideration the winning condition and the raw score.

\subsubsection{Open-Loop Expectimax Tree Search (OLETS)} It is based on the Hierarchical Open-Loop Optimistic Planning (HOLOP), improved to work better in stochastic environments. Its details and the differences between both algorithms are described in \cite{perez20162014}. The version provided by the framework has a playout length of $5$ and the default value function takes into consideration the winning condition and the raw score.

\subsubsection{Rolling Horizon Evolutionary Algorithm (RHEA)} It is an evolutionary algorithm that uses a population of individuals, which represents a sequence of actions to execute in a specific order. The plan of actions of each individual is evaluated using a forward model and the agent takes the first action of the individual with the best fitness \cite{gaina2017analysis}. The \textit{sampleRHEA} provided in the framework uses the \textit{WinScoreHeuristic}, evolves the individuals keeping $10$ at a time and has a mutation rate of $1$. It applies mutation and crossover to get the next generation until the time runs out. Because it produces as many sequences as possible in this time, the number of individuals is dynamic.

\subsubsection{Random Search (RS)} It is similar to RHEA, but it randomly generates the individuals instead of evolving them. The \textit{sampleRS} uses lengths of $10$, and it produces as many sequences as possible in the given time.
\\\\
The algorithms that we executed in the experiments as a point of comparison were not modified in any way. However, those that we included as part of the EDS required some modifications that do not affect their core implementation but allow them to fit the needs of the system. These include those related to the abstraction of the heuristics \cite{guerrero2017beyond}, and the modifications needed to make the algorithms return Opinions instead of the action to take. As detailed in the next section, an Opinion is a simple data structure which holds the action to take, a value computed for that action and the name of the algorithm which suggested it.

\subsection{Game Heuristics}

The EDS combines different algorithms and game heuristics. The heuristics that we utilize go further than merely winning the game by following score, and are based on the work of Guerrero-Romero \textit{et al.} \cite{guerrero2017beyond}.

\subsubsection{Winning Maximization Heuristic (WMH)}
Its goal is winning the game, maximizing the score when reaching the winning status is not possible.

\subsubsection{Exploration Maximization Heuristic (EMH)}
Its goal is maximizing the exploration of the level, prioritizing visiting those locations that were not visited before, or have been visited less often. For these experiments, this heuristic has been updated regarding the one in \cite{guerrero2017beyond}, so winning the game is always rewarded instead of penalized.

\subsubsection{Knowledge Discovery Heuristic (KDH)}
Its goal is interacting with the game as much as possible to trigger interactions, and new sprite spawns. For these experiments, this heuristic has been updated regarding the one in \cite{guerrero2017beyond}, so winning the game is always rewarded instead of penalized.

\subsubsection{Knowledge Estimation Heuristic (KEH)}
Its goal is interacting with the game to predict the outcomes of the interactions between the different elements of the game, related to both the victory status and score modifications.

\section{Ensemble Decision Systems}\label{sec:EDS}
Each of the algorithms described have their own strengths and weaknesses. Each algorithm can solve a different set of problems, but none can currently solve all of them.

Ensemble Decision Systems are a flexible system for combining the decisions of multiple algorithms into a single action. Instead of trying to develop complex problem solving behaviour in a single algorithm, complexity is built up with simpler layers of behaviours that each focus on different aspects of a problem, or different types of problems altogether.

Expanding the capabilities of an algorithm whilst maintaining the strengths it already has can be challenging. EDS's offer a potential solution to this by allowing each algorithm to focus on its strengths, and addressing their weaknesses through other algorithms. Intuitively, if an algorithm is good at finding paths through an environment, but bad at identifying long term goals, then combining that initial algorithm with a long term planner may give the overall system the best of both worlds.

\subsection{Architecture}
The EDS is comprised of two components. The algorithms, known as \emph{voices}, which evaluate the current state, and the \emph{arbitrator} (Section \ref{sec:arbitrator}), which makes the final decision of which action to take. When a game begins, the arbitrator is given the state which is then passed to each voice. The voices perform their own analysis and then return their \emph{opinion} to the arbitrator. An opinion is a data structure which holds the action selected, as well as a value assigned to that action.

Once every voice has returned their opinion, the arbitrator will use a final action selection policy to decide which action to take, based on the information from the voices.

This architecture can be adjusted in a variety of ways. First of all, the number of voices, and which are used, can be altered at implementation or at run-time. For example, if a voice does not seem to be performing well in a given game, it can be disabled or swapped with another voice to improve overall system performance. The action selection policy is also a parameter which can be adjusted. Some examples would be using a bandit selection algorithm to decide which action to take from the opinions, or a diplomatic option which selects the action that most voices have selected. A neural network could also be trained to identify which voices do well in certain states, as an action selection policy. Each voice could potentially assign a value to all possible actions, and the action selection policy then decides based on the highest value, and could potentially decide actions that none of the voices had primarily suggested. There are a wide number of possibilities available for adjusting the EDS.

\subsection{Action Selection Policies}
The policy used to select the action to take has a large impact on the overall behaviour of the EDS. The variations that were used for the experiments are the following:

\subsubsection{Highest Value}
This policy simply selects the action which returns the highest value, based on the analysis of its voice. The possible range from this is not currently limited, though future work would look at normalizing the output from the heuristics.

\subsubsection{Diplomatic}
This policy is only useful when using more than two voices in the system, and essentially selects the action that has the most votes from the voices. The action that is returned with each Opinion is counted as a vote in favour of that action. In the case that no majority exists, then an alternative action selection policy is used. For the experiments in this work, this was the random action selection policy.

\subsubsection{Random}
This policy simply selects an opinion at random and uses the action assigned to it.

\subsection{Arbitrator} \label{sec:arbitrator}
The arbitrator can be implemented in a number of different ways, depending on the constraints of the problem set. In particular, the GVGAI has a $40$ms per action time limit that needs to be respected, which has a significant impact on the analysis that algorithms can do. This has a large impact on an EDS because each algorithm has to be allocated a slice of that $40$ms. Further impacting this, the GVGAI does not allow multi-threading, which is where an EDS would be able to leverage the voices simultaneously.

In order to deal with these limitations, two different arbitrators, described in this section, were created and tested.

\subsubsection{Central Arbitrator}
This arbitrator splits the $40$ms evenly between all of the voices within the system. As an example, if there are N voices, then each voice receives $(40/N)-1$ ms per time-step to return its opinion. 

\subsubsection{Asynchronous Arbitrator}
This arbitrator gives each of the voices within the system the full $40$ms decision time, but does so by skipping actions until each voice has returned an opinion. If there are N voices within the system, then for N-$1$ time-steps a Nil action is returned and a voice performs its analysis. At the Nth time-step, after all voices have returned their Opinions, then an action is selected based on the current action selection policy. The decision of taking no-op actions comes from the will to affect the game the least, so the voices analyze the most similar state of the game as possible.

\subsection{Variations}
Some of the variations that were tested for this work are outlined in the following sections.

\subsubsection{BestFour}
% The best four algorithms for each of the heuristics using a highest value action selection policy
This variation makes use of prior work, which measured the performance differences in playing GVGAI games with different heuristics \cite{guerrero2017beyond}. The heuristics created for that work with the algorithms identified as performing best for each heuristic were used here in an EDS. There are four voices in this variation (OLETS with WMH, RS with EMH, RS with KDH and OLETS with KEH), the Central Arbitrator is used, and the action selection policy is Highest Value.

\subsubsection{BestFourDiplo}
% The best four algorithms for each of the heuristics using a diplomatic action selection policy
This variation uses the same four voices as BestFour (OLETS with WMH, RS with EMH, RS with KDH and OLETS with KEH) and the Central Arbitrator, but it uses Diplomatic as the action selection policy instead.

\subsubsection{BestExpSc}
% The best algorithms for MaxExploration and MaxScore using a highest value action selection policy - PS: Diplomatic doesn't make any sense in the case of just two voices
There are two voices in this variation: OLETS with WMH and RS with EMH. The Central Arbitrator is used, and the action selection policy is Highest Value.

\subsubsection{MCTSExpSc}
% MCTS MaxExploration and MaxScore with highest value action selection
There are two voices used in this variation: MCTS with WMH and MCTS with EMH. This variation was tested to give a more direct comparison to the sampleMCTS controller. The Central Arbitrator is used and the action selection policy is Highest Value.

\subsubsection{OLETSExpSc}
% OLETS MaxExploration and MaxScore with highest value action selection
There are two voices used in this variation: OLETS with WMH and OLETS with EMH. This variation was tested to give a more direct comparison to the OLETS controller. The Central Arbitrator is used, and the action selection policy is Highest Value.

\subsubsection{OLETSExpScAsync}
% OLETS MaxExploration and MaxScore with highest value action selection with Async Arbitrator
There are two voices used in this variation: OLETS with WMH and OLETS with EMH. It uses the the Asynchronous Arbitrator and Highest Value as the selection policy. This variation was also tested to give a more direct comparison to the OLETS controller.

\section{Game Selection}
% Why were certain games chosen? Which are deceptive and how do their performance measures work? 
There are $122$ available single-player games in the GVGAI Framework at the time of writing\footnote{\url{https://github.com/GAIGResearch/GVGAI/blob/master/examples/all_games_sp.csv}} and running the experiments using all of them is prohibitively time consuming. Therefore, a subset of these was selected, with enough diversity to ensure that the games selected represent the full variety of options available from the GVGAI set.

The choice of games for the experiment was based on previous work. Twenty games were based on Gaina \textit{et al.} selection in \cite{gaina2017population}, where they used a balanced set of $20$ stochastic and deterministic games. We also included the ten games which collectively provided the highest information gain in Stephenson \textit{et al.} work in \cite{stephenson2018continuous}. Because $2$ of the games are present in both sets (\textit{Chopper} and \textit{Escape}), the total number of games used in the experiment was $28$ (Table~\ref{table:games}).

\begin{table}[htbp]
\caption{The Games selected from the GVGAI Framework (games in bold are deceptive)}
\begin{center}
\begin{tabular}{|c|c|c|c|}
\hline
\textit{Aliens} & \textit{Avoidgeorge} & \textit{Bait} & \textit{\textbf{Butterflies}}\\
\hline 
\textit{CamelRace} & \textit{Chase} & \textit{Chopper} & \textit{Crossfire}\\
\hline
\textit{Digdug} & \textit{Escape} & \textit{Freeway} & \textit{Hungrybirds}\\
\hline
\textit{Infection} & \textit{Intersection} & \textit{\textbf{Invest}} & \textit{Labyrinthdual}\\
\hline
\textit{\textbf{Lemmings}} & \textit{Missilecommand} & \textit{Modality} & \textit{Plaqueattack}\\
\hline
\textit{Roguelike} & \textit{Seaquest} & \textit{\textbf{Sistersavior}} & \textit{Survivezombies}\\
\hline
 \textit{Tercio} & \textit{Waitforbreakfast} & \textit{Watergame} & \textit{Whackamole}\\
\hline
\end{tabular}
\label{table:games}
\end{center}
\end{table}
% Full List of Games:
% Aliens, Avoidgeorge, Bait, CamelRace, Chase, Chopper, Crossfire
% Digdug, Escape, Freeway, Hungrybirds, Infection, Intersection
% Invest, Labyrinthdual, Lemmings, Missilecommand, Modality, Plaqueattack
% Roguelike, Seaquest, Sistersavior, Survivezombies, Tercio
% Waitforbreakfast, Watergame, Whackamole
% Freeway, Invest, Labyrinthdual, Tercio, Sistersavior, Avoidgeorge, Escape, Whackamole, Chopper and Watergame were selected based on work here https://arxiv.org/abs/1809.02904
%\cite{stephenson2018continuous}
% The selection of games in the experiments I carried out were based in the ones selected for the work done in here: https://ieeexplore.ieee.org/abstract/document/7969540
% Second relevant paper for selection
%\cite{gaina2017population}

The selection includes a series of deceptive games, which are designed to lead the agent away from the most optimal solution \cite{anderson2018deceptive}:

\paragraph{\textit{Butterflies}} The goal is capturing all of the butterflies before the time runs out or all the cocoons open. A cocoon opens when a butterfly collides with it, spawning a new butterfly. Each butterfly captured increases the score $+2$. Therefore, the higher the number of cocoons opened, the higher the final score can be.  Trying to win quickly leads to a less optimal solution, as the score will not be as high as it could.

\paragraph{\textit{Invest}} The goal is maximizing the score by collecting gold coins and investing them with the investors. Investing too much and getting into debt results in a loss. Once an investment is made, that investor will disappear for a set period, before returning and awarding a higher amount of points to the player. Each gold coin awards $1$ points when collected; the Green investor takes $3$ points and returns $5$ after $30$ timesteps;
the Red investor takes $7$ points and returns $15$ after $60$ timesteps; the Blue investor takes $5$ and returns $10$ after $90$ timesteps.

\paragraph{\textit{Lemmings}} Lemmings are spawned from one door and try to get to the exit of the level. There are obstacles on the way, so the goal is destroying these so the lemmings can reach the exit. For every lemming that reaches the exit $+2$ is given, but $-1$ is subtracted from every piece of wall destroyed. If a lemming falls into a trap, the score is reduced $-2$, and if it is the avatar who falls, the score is reduced $-5$.

\paragraph{\textit{Sistersavior}} The goal is rescuing all of the hostages to be able to defeat the scorpion and win the game. There is the option to kill the hostages, receiving a moderate immediate reward ($+2$ points if shot vs. $+1$ if rescued). The scorpion provides $14$ points if defeated, but it can only be defeated if all $3$ hostages are rescued.

\section{Experimental Work}
This section describes the experiments that were carried out and provides instructions for replication.

\subsection{Experimental Setup and Specifications}\label{sec:specs}
The code used for the experiments is in a Github repository\footnote{https://github.com/Damorin/PhD-Ensemble-GVGAI/tree/master/src/COGPaper}. All of the experiments were performed using a laptop running Linux Mint with an Intel i7-8565U CPU with 16GB of RAM.

The GVGAI framework has up to five levels available for each of the games. Each level was run $30$ times in order to build up confidence in the results being gathered, having $150$ runs per game for each controller. Each successful iteration of a game that resulted in either a win or loss without any crashes produces one unit of data, containing the information $\lbrack agent, game, score, win/lose, time\rbrack$. 
%This represents a single unit of data. 
In total, $48,600$ units of data were generated from these experiments, with each game generating $1800$ units of data across all of the agents.

The controllers described in Sections~\ref{Controllers} and~\ref{sec:EDS} were used for these experiments. The final list of controllers can be found in Table~\ref{table:controllers}.

\begin{table}[htbp]
\caption{Full list of controllers used}
\begin{center}
\begin{tabular}{|l|l|}
\hline
\textbf{EDS Agents} & \textbf{Sample Agents}                \\ \hline
BestExpSc       & OLETS                       \\ \hline
BestFourDiplo                 & sampleMCTS                  \\ \hline
BestFour               & sampleOneStepLookAhead      \\ \hline
MCTSExpSc       & sampleRandom                \\ \hline
OLETSExpSc      & sampleRHEA                  \\ \hline
OLETSExpScAsync & sampleRS                    \\ \hline
\end{tabular}
\end{center}
\label{table:controllers}
\end{table}

\subsection{Results}

%To analyze the results of the experiments, we have considered two different metrics: the total number of wins and the average score
To thoroughly analyze the performance of the GVGAI agents, both wins and scores are taken into account. A number of visualizations and a ranking of performance based on those metrics have been created. These are detailed below:
%Not all games can be played optimally by simply winning, particularly the deceptive games which often use an ambiguous reward structure to lure agents away from optimal performance.

\subsubsection{Percentage of wins per game}

\begin{figure}[htbp]
\centerline{\includegraphics[width=9cm,keepaspectratio]{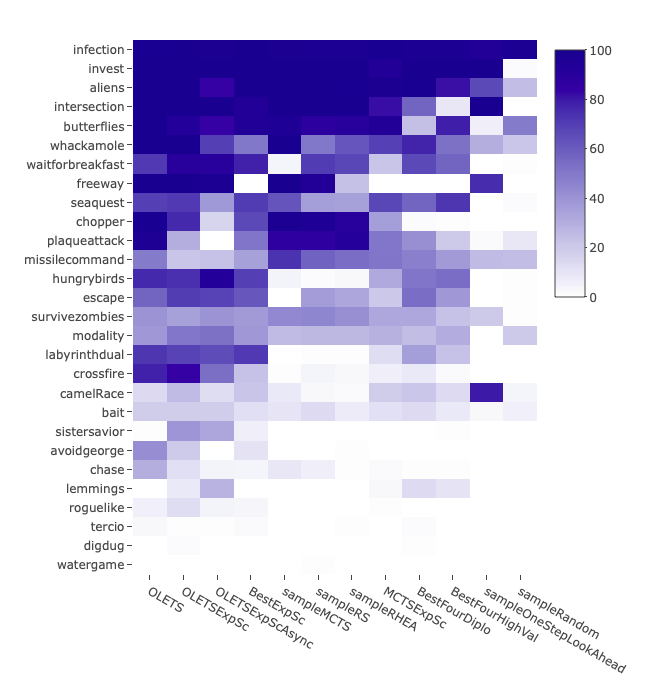}}
\caption{Total percentage of wins ($0-100\%$) of each controller by game.}
\label{fig:results_wins}
\end{figure}

Fig.~\ref{fig:results_wins} shows the percentage of wins that each agent achieved in each game. It has been ordered to show the win rates per game, with the easier games to win at the top and the harder at the bottom.

\subsubsection{Score average per game}

\begin{figure}[htbp]
\centerline{\includegraphics[width=9cm,keepaspectratio]{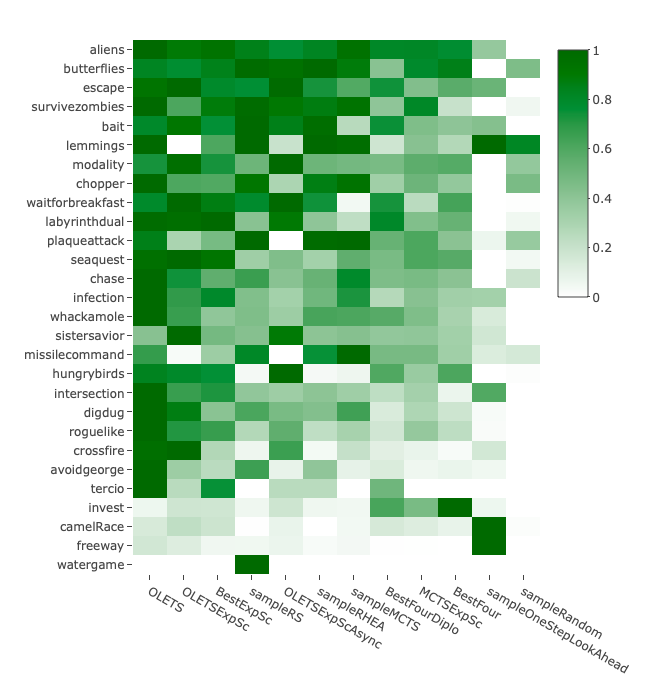}}
\caption{Average scores per game. To be able to make a comparison between the different games, the scores have been normalized, taking the maximum and minimum average scores obtained in each game as limits for that game.}
\label{fig:results_score}
\end{figure}

Fig.~\ref{fig:results_score} shows the average score that each agent achieved per game. The score depends on the game considered, so to have all the results on the same scale to be able to be displayed together, the average of the scores per game were normalized. This normalization was done by taking the maximum and minimum score average obtained by the agents in a game, and used those values for the normalization, so that for every game, there is an agent whose score is represented by $0$ and another represented by $1$, and all others in between. This approach visualizes the difference in scores between the agents, by game, in an informative manner. It is possible to distinguish between those games were agents manage to get a score easily, and those were they struggle, and the contrast between the higher and lower average obtained.

\subsubsection{F1 Ranking}
In order to have an overall performance of the agents through every game, we include a Formula 1 (F1) ranking, like the one used in the GVGAI Competition \cite{perez20162014}. Each agent receives $25$, $18$, $15$, $12$, $10$, $8$, $6$, $4$, $2$, $1$ or $0$ points per game based on their performance regarding the rest of the algorithms in that game. The performance is measured with the data that was gathered at the end of each run (Section~\ref{sec:specs}), considering the total number of wins, the average of the score and the average of timesteps per game. In general, the agent with a higher number of wins is better in that game. In case there is a tie, the highest average score between the agents breaks the tie. In the case that there is a tie in both measures, the lowest average of timesteps is considered. The highest ranked agent receives $25$ points in that game, the second one $18$, and so on, until all agents have been assigned a score. If there is a tie in the three measures, those agents involved in it will receive the same number of points. Table~\ref{table:ranking} shows the final ranking overall the $28$ games used in the experiment. 

\begin{table}[htbp]
\caption{Controllers ranked by the total number of F1-points gained overall the games. It includes the total percentage of wins (\%Wins) and the number of different unique games won (\#Games) by each of them.}
\begin{center}
\begin{tabular}{|c|c|c|c|c|}
\hline
\textbf{} & \textbf{Controller} & \textbf{F-1} & \textbf{\%Wins} & \textbf{\#Games}\\
\hline
\textbf{1} & OLETS & \cellcolor{blue!15}$501$ & \cellcolor{blue!15}$55.69\%$ & $25$\\
\hline
\textbf{2} & OLETSExpSc & $463$ & $53.59\%$ & \cellcolor{blue!15}$27$\\
\hline
\textbf{3} & BestExpSc & $318$ & $42.86\%$ & $24$\\
\hline
\textbf{4} & OLETSExpScAsync & $316$ & $45.52\%$ & $24$\\
\hline
\textbf{5} & sampleMCTS & $253$ & $40.00\%$ & $19$\\
\hline
\textbf{6} & sampleRS & $237$ & $39.76\%$ & $22$\\
\hline
\textbf{7} & BestFourDiplo & $213$ & $33.02\%$ & $23$\\
\hline
\textbf{8} & sampleRHEA & $194$ & $36.71\%$ & $23$\\
\hline
\textbf{9} & BestFour & $182$ & $28.93\%$ & $22$\\
\hline
\textbf{10} & MCTSExpSc & $181$ & $33.29\%$ & $22$\\
\hline
\textbf{11} & sampleOneStepLookAhead & $88$ & $21.48\%$ & $12$\\
\hline
\textbf{12} & sampleRandom & $29$ & $9.29\%$ & $15$\\
\hline
\end{tabular}
\label{table:ranking}
\end{center}
\end{table}

\subsection{Discussion}\label{sec:discussion}
%Not all games can be played optimally by just winning, particularly the deceptive games which often use an ambiguous reward structure to lure agents away from optimal performance.
The first finding is that overall the best performing agent is the OLETS agent, which gets a final score of $501$ points and wins the most amount of games out of all of the agents ($55.69\%$). The second highest, OLETSExpSc, has a similar level of performance ($53.59\%$), though the games that it wins are quite different from OLETS. In particular, it is able to outperform OLETS on a variety of games, such as \textit{Waitforbreakfast}, \textit{Crossfire} and \textit{Escape}. OLETSExpSc also appears to make progress towards winning over a wider range of games than OLETS, with OLETS winning at least once $25$ unique games, as opposed to OLETSExpSc, winning $27$. 

In regards to deceptive games, which are currently not solved by the sample agents, such as \emph{SisterSavior} and \emph{Lemmings}, EDS agents manage to achieve significant results. Interestingly, the best-performing agents on these levels are EDS agents which use OLETS as their voices, but the standard OLETS algorithm is not able to progress on these games. This improvement in the OLETS performance suggests that combining the traditional OLETS algorithm with a dedicated search algorithm provides enough of an advantage to allow OLETS to solve more problems beyond its original capability.

OLETSExpSc gets $59$ of $150$ wins in \textit{SisterSavior}, OLETSExpScAsync, $50$, and BestExpSc, $9$, when OLETS gets $1$ and the rest of the algorithms, none. In particular, the EDS algorithms are able to consider compromising their immediate score reward (saving the hosts instead of killing them), having the chance to reach the most challenging goal and win the game.

\begin{figure}[htbp]
\centerline{\includegraphics[width=9cm,keepaspectratio]{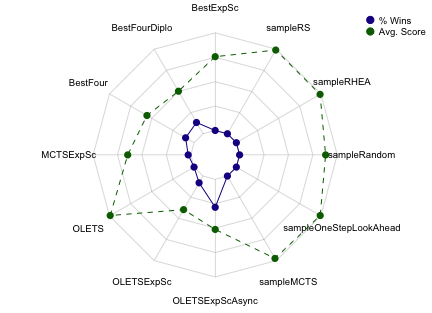}}
\caption{\textit{Lemmings} stats per agent. The pct. of wins is in range [$0, 100\%$] and the avg. of score is relative to the maximum and minimum scores achieved in the game (range: [$-143, 1$]). Values have been normalized in this figure.}
\label{fig:sp_lemmings}
\end{figure}

In \textit{Lemmings}, the EDS agents that manage to win the game are MCTSExpSc, OLETSExpSc, BestFour, BestFourDiplo and OLETSExpScAsync, with $4$, $12$, $15$, $20$ and $43$ of $150$ wins respectively. As Fig~\ref{fig:sp_lemmings} shows, they also have the lowest score average, as they use heuristics with further considerations other than trying to maximize the score, not being afraid of losing points while exploring the level and destroying the walls. Even when the solutions might not be the most optimal ones, and their levels could be solved getting a higher score average, they are able to win the game when the other controllers fail.

\begin{figure}[htbp]
\centerline{\includegraphics[width=9cm,keepaspectratio]{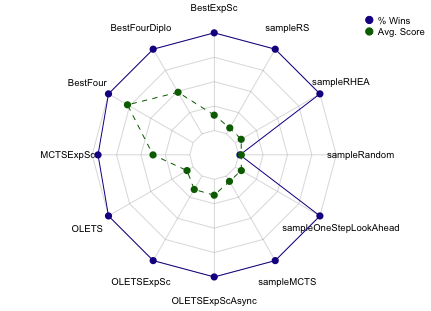}}
\caption{\textit{Invest} stats per agent. The percentage of wins is in range [$0, 100\%$] and the avg. of score is relative to the maximum and minimum scores achieved in the game (range: [$-7, 161$]). Values have been normalized in this figure.}
\label{fig:sp_invest}
\end{figure}

Looking further at the deceptive games, \emph{Invest} has some interesting results (Fig.~\ref{fig:sp_invest}). All of the sample agents score an average of five or less across their games of \textit{Invest}, while the EDS agents manage to score significantly higher. This is an interesting result as the deception in \textit{Invest} was designed to trick search based agents that would typically be reluctant to surrender their current score in order to progress. The EDS agents are able to perform well in this game, with BestFour scoring highest with an average of $123$. Interestingly, one of the EDS agents, MCTSExpSc, loses some of the games. As the sampleMCTS agent does not lose any games of \textit{Invest} in the experiments, it seems that the exploration combined with MCTS causes it to take some risks, while increasing its overall performance.

Games where OLETS still performs notably better than its EDS versions, are discussed next. 
In \textit{Missilecommand}, OLETSExpSc gets $22.00\%$ of wins and  OLETSExpScAsync $22.67\%$, in contrast with OLETS's $48.67\%$. We believe that this decrease is due to the exploratory behaviour added to these agents, which brings them to explore the level instead of focusing on achieving the goal of the game. When the agents finally realize they are about to lose, they are too far away to be able to prevent themselves from losing the game. 
In \textit{Chopper}, OLETS manages to win most of the games ($99.33\%$ of wins) with an average of score of $17.03$. In contrast, OLETSExpSc achieves a $76.00\%$ of wins and an average of score of $4.5$, and OLETSExpScAsync gets a $16.00\%$ of wins averaging $-5.55$ points. In \textit{Plaqueattack}, the decrease of performance is even higher: OLETS manages to get $96.00\%$ of wins when OLETSExpSc only wins a $30.00\%$ of the games and OLETSExpScAsync does not win any of them. Looking at the behaviour of the agents in both of these games, during observation the Async controller spends most of the time exploring rather than learning to gain points. The central controller moves much faster, reaching new positions sooner and, therefore, lessening the value of the exploration heuristic sooner. It allows it to focus on the score and win the game. This may be due to the exploration heuristic providing excessively high evaluations in the decision process of these agents, causing them to ignore the actual goal. Hence, the value returned by the EDS agents' heuristics should be tuned in future work to have a more balanced input from all of the heuristics in levels with a large state space, particularly where the rewards are focused in a small area of the level.
%Hence, in order to have a more balanced input from both of the heuristics in the EDS agents in levels with a large state space, particularly where the rewards are focused in a small area of the level, the value returned by the heuristics should be tuned in future work.

\textit{CamelRace} is a racing game where the avatar must get to the finish line before any other camel does. It is a compelling case to comment on, as it is quite a simple game that the complex algorithms struggle to solve but OSLA, one of the simplest controllers in the sample batch, gets $120$ of $150$ wins. Some EDS agents perform better than OLETS ($21$ wins) in this game (MCTSExpSc, BestFourDiplo, BestExpSc and OLETSExpSc, with $28$, $32$, $33$ and $38$ wins respectively) but they are far behind OSLA. The reason for the good performance of OSLA comes from how fast the game is, the well defined path that should be followed and the lack of score rewards until the end. % The game is lost as soon as another camel reaches the goal and, unless the agents are heading to the exit, they will lose. 
%OSLA's heuristic (SimpleStateHeuristic) considers winning condition, score and the positions to other objects of the game. 
Because the win state is not reached unless the agent heads to the goal, and there is no score information to use, the agents that just use these values as a reference will behave randomly until their movement heads them to the exit. If those agents are lucky enough to choose those actions that get them closer to it quickly, then they may win. Exploring in this game is also not beneficial, because the path to the exit should be followed quickly to avoid losing. 
OSLA's heuristic (SimpleStateHeuristic) considers, apart from winning condition and score, those actions where the distance between the avatar and the closest portal sprite (which is the category of the goal sprite) is small. In the first four levels, where this information is enough to reach the goal, it manages to get there promptly. However, in the last level, where there is a wall between the agent and the goal, it gets stuck not being able to rectify its path.

\begin{figure}[htbp]
\centerline{\includegraphics[width=9cm,keepaspectratio]{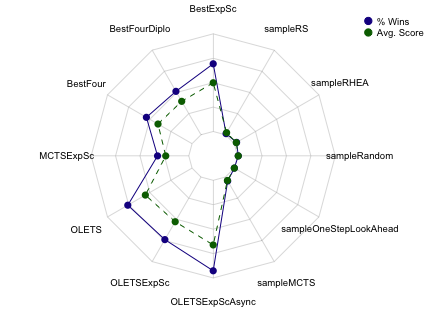}}
\caption{\textit{Hungrybirds} stats per agent. The pct. of wins is in range [$0, 100\%$] and the avg. of score is relative to the maximum and minimum scores achieved in the game (range: [$0, 140$]). Values have been normalized in this figure.}
\label{fig:sp_hungrybirds}
\end{figure}

Finally, an example where the exploration heuristic is profitable is \textit{Hungrybirds}. In this game, the player is a bird that needs to exit a maze but needs to find food and eat it while doing so to avoid dying. %The rate of wins and the average of the score are highly related in this game as points are received when food or the goal are reached ($40$ and $100$ points respectively) and food it is requested to win the game. 
As observed in Fig.~\ref{fig:sp_hungrybirds}, the only sample agent with a significant number of wins is OLETS, while the rest of the sample controllers achieve less than a $4.60\%$ winning rate. In contrast, all EDS agents based on sample controllers achieve between $32.00\%$ and $74.00\%$ win rate, and OLETSExpScAsync improves OLETS ($92.67\%$ vs $76.00\%$). Something similar happens in other games such as \textit{Escape}, \textit{Labyrinthdual} and \textit{Lemmings}.
%in contrast with all of the EDS agents, even when they use other controllers in their systems apart from it. 
%These results show how using agents with different policies can help to solve a game, showing the potential of the approach developed.
These results show how using agents with different policies can help to improve the win rates in a game, showing the potential of the approach developed.

\section{Conclusions and Future Work}
This work presents a novel approach to designing GVGP agents through the use of an Ensemble Decision System (EDS) and showcases a number of potential variations of how such a system may work. While the EDS does not currently outperform the individual algorithms, it maintains a significant portion of their strengths and adds to the generality of the agent by allowing it to succeed at more games. The flexibility of the EDS is the main strength that it offers, and the potential of this approach has not yet been fully explored.

The results show the potential of combining different algorithms with different goals using this kind of system. OLETS' EDS variant is able to win two more games than the original agent without compromising its overall performance, simply by providing two different intentions to take into consideration. One of these games, \textit{Lemmings}, is a deceptive game that none of the sample controllers are able to solve but five different EDS systems win, one of them $43$ times. This is an impressive result that shows what this approach is capable of achieving, having noted potential that could be developed further. A similar generality improvement is seen between the sampleMCTS and MCTSExpSc, winning in $19$, vs $22$ games.

Future work for this project will look at developing further variations of the system, which could look at different action selection policies or types of arbitration. Possibilities in this area could include developing a bandit style selection algorithm for deciding which voice should be selected, or the arbitrator itself could be a neural network that is trained to identify which voices are best suited to deal with a given state. It will also compare EDS to portfolio approaches as they have distinct strengths and weaknesses that would be interesting to analyze in detail.

At the moment the EDS variations that have been implemented are quite naive. There has not been any optimization of which voices work best together, or development of the more sophisticated arbitrators and action selection policies. Each voice is treated as an expert, and their evaluation is trusted. This creates some performance issues for certain games, that were highlighted earlier in Section~\ref{sec:discussion}. In particular, it seems that there are cases where the exploration heuristic might be providing more weight in the decision process of the agents than it should. This has been noticed especially in games that present significant exploratory areas but where the primary objective of the game is focused on a particular area. %Further research is also needed to tune the heuristic values and balance the voices to have equal input independently of the type of game considering. 
%Normalizing the voices so they are on the same scale is not that easy, as, for example, the minimum and maximum scores achievable in a game are unknown to the agent, which makes it challenging to develop a balanced approach that would regard the score and exploration with the same \textit{level of measure} during gameplay. 
This problem could be improved by not only tuning the values returned by the heuristics, but by adding in voices that are designed to fulfill a specific purpose. As an example, a survival voice could be added to an EDS whose role would either be to veto actions that lead to a losing state or remove an action from consideration. Or perhaps all voices could be able to veto actions that they evaluate as being dangerous. A fitness function could be built into the arbitrator that is able to evaluate how well a voice is performing in a particular situation, and could adjust the confidence of that voice or swap it with another voice. Voices themselves could be required to evaluate all actions available, and the system could then make selections based on evaluations from all perspectives. Lastly, a voice does not need to be a singular algorithm, but could instead be a collection of algorithms working together to return a single opinion. An EDS of multiple EDSs. %The arbitrator could similarly be an entirely different mechanism, such as a neural network or a bandit style function, selecting voices which have shown themselves previously to be effective in similar states.

In short, the EDS offers a considerable amount of flexibility and consolidates the strengths of current GVGP approaches to overcome their weaknesses, making the system more robust overall. It is also worth noting that while a range of variations were built for these experiments, they are still naive in their implementation. Further sophistication in their decision making process, and a more intelligent selection process regarding which voices to pair together will likely improve performance further. With the recent trend of high performing solutions coming not from singular algorithms, but from algorithms working together, the interesting results of this paper show that this is a fruitful area for further research.

\section*{Acknowledgment}

This work was partially funded by the EPSRC CDT in Intelligent Games and Game Intelligence (IGGI) EP/L015846/1.

%--------------------------------------
% REFERENCES
%-------------------------------------

\bibliographystyle{IEEEtran}
\bibliography{references}

\end{document}